\documentclass[letterpaper]{article} 
\usepackage{aaai19}
\usepackage{times}
\usepackage{helvet}
\usepackage{courier}
\usepackage{url}
\usepackage{graphicx}
\usepackage{latexsym}
\usepackage{soul}
\usepackage{color}
\usepackage{float}
\usepackage{tabularx}
\usepackage{subfig}
\usepackage{multirow}
\usepackage{latexsym}
\usepackage{amsmath}
\usepackage{amssymb}
\usepackage{colortbl}
\usepackage{array}
\usepackage{bm}
\usepackage{booktabs}
\usepackage{arydshln}
\usepackage{pgfplots}

\usepackage{times}
\usepackage{xcolor}
\usepackage{subfig}
\usepackage{subfloat}
\usepackage{tikz-dependency}
\usepackage{algorithm}
\usepackage{algorithmic}
\usepackage{tabularx}
\usepackage{xparse}
\usepackage{xspace}
\usepackage{relsize}
\usepackage{graphicx}
\usepackage{multirow}
\usepackage[normalem]{ulem}
\usepackage{amsmath}
\usepackage{booktabs}
\usepackage{CJKutf8}

\pgfplotsset{every axis/.append style={
                    axis x line=middle,    
                    axis y line=middle,    
                    axis line style={->}, 
                    xlabel={$x$},          
                    ylabel={$y$},          
                    label style={font=\tiny},
                    tick label style={font=\tiny},
                    legend style={font=\tiny},
                    legend pos=outer north east
                    }}

\tikzset{>=stealth}

\def\ba{\mathbf{a}}
\def\bb{\mathbf{b}}

\def\be{\mathbf{e}}

\def\bh{\mathbf{h}}

\def\bs{\mathbf{s}}

\def\bx{\mathbf{x}}

\def\bE{\mathbf{E}}

\def\bR{\mathbf{R}}

\def\bW{\mathbf{W}}

\def\cJ{\mathcal{J}}

\def\cL{\mathcal{L}}

\newcommand{\citet}[1]
{\citeauthor{#1}~\shortcite{#1}}
\newcommand{\citep}{\cite}

\DeclareMathOperator*{\argmax}{arg\,max}
\DeclareMathOperator*{\softmax}{softmax}
\DeclareMathOperator*{\LSTM}{LSTM}
\DeclareMathOperator*{\BiLSTM}{Bi-LSTMs}

\DeclareMathOperator*{\Switch}{Switch}
\DeclareMathOperator*{\SwitchLSTM}{Switch-LSTMs}
\DeclareMathOperator*{\BiSwitchLSTM}{Bi-Switch-LSTMs}

\newcommand{\tabincell}[2]{\begin{tabular}{@{}#1@{}}#2\end{tabular}}

    \def\addlegendimage{\csname pgfplots@addlegendimage\endcsname}

\pgfplotsset{
cycle list={%
{draw=black,mark=star,solid},
{draw=black, mark=square,solid}}}

\newenvironment{itemize*}%
  {\begin{itemize}%
    \setlength{\itemsep}{0pt}%
    \setlength{\parskip}{0pt}}%
  {\end{itemize}}
  \newenvironment{enumerate*}%
  {\begin{enumerate}%
    \setlength{\itemsep}{0pt}%
    \setlength{\parskip}{0pt}}%
  {\end{enumerate}}

\makeatletter
\newcommand{\printfnsymbol}[1]{%
  \textsuperscript{\@fnsymbol{#1}}%
}
\makeatother

\title{Switch-LSTMs for Multi-Criteria Chinese Word Segmentation}

\author{Jingjing Gong\thanks{\hspace{1mm} J. Gong and X. Chen contributed equally to this paper.}, Xinchi Chen\printfnsymbol{1}, Tao Gui, Xipeng Qiu\thanks{\hspace{1mm} Corresponding author.}\\
Shanghai Key Laboratory of Intelligent Information Processing, Fudan University \\
School of Computer Science, Fudan University\\
Shanghai Insitute of Intelligent Electroics \& Systems \\
\{jjgong, xinchichen13, tgui16, xpqiu\}@fudan.edu.cn
}

\date{}

\begin{document}
\maketitle
\begin{abstract}
  Multi-criteria Chinese word segmentation is a promising but challenging task, which exploits several different segmentation criteria and mines their common underlying knowledge. In this paper, we propose a flexible multi-criteria learning for Chinese word segmentation. Usually, a segmentation criterion could be decomposed into multiple sub-criteria, which are shareable with other segmentation criteria. The process of word segmentation is a routing among these sub-criteria. From this perspective, we present Switch-LSTMs to segment words, which consist of several long short-term memory neural networks (LSTM), and a switcher to automatically switch the routing among these LSTMs. With these auto-switched LSTMs, our model provides a more flexible solution for multi-criteria CWS, which is also easy to transfer the learned knowledge to new criteria.
  Experiments show that our model obtains significant improvements on eight corpora with heterogeneous segmentation criteria, compared to the previous method and single-criterion learning.
\end{abstract}

\begin{CJK*}{UTF8}{gbsn}

\begin{figure*}
  \centering
  \subfloat[Single criterion CWS]{
  \includegraphics[width=0.11\textwidth]{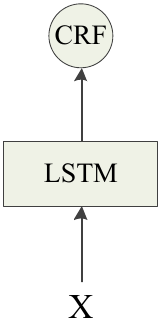} \label{fig:Model-I}
  }
  \hspace{3em}
  \subfloat[Multi-criteria CWS with MTL]{
  \includegraphics[width=0.34\textwidth]{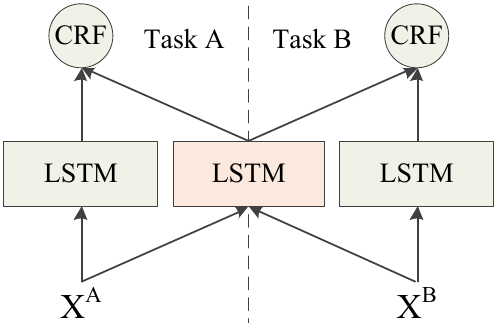} \label{fig:Model-II}
  }
  \hspace{3em}
  \subfloat[Multi-criteria CWS with Switch-LSTMs]{
  \includegraphics[width=0.158\textwidth]{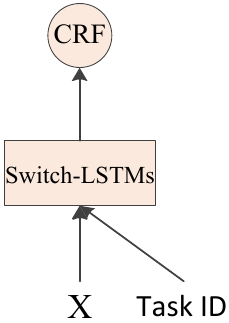} \label{fig:Model-III}
  }
  \caption{Architectures of single-criterion and multi-criteria Chinese word segmentation. Parameters of components in orange are shared, and those in green are private.}\label{fig:cws_models}
\end{figure*}

\section{Introduction}
Chinese word segmentation (CWS) is a preliminary and important task for Chinese natural language processing (NLP). Currently, the state-of-the-art CWS methods are based on supervised machine learning algorithms and greatly rely on a large-scale annotated corpus whose cost is extremely expensive.
Although several CWS corpora have been built with great efforts, their segmentation criteria are different. Since these segmentation criteria are from different linguistic perspectives, it remains to be a challenging problem on how to effectively utilize these resources.

Most existing methods focus on improving the performance of each single segmentation criterion. Recently, deep learning methods have been widely used in segmenting Chinese words and can effectively reduce the efforts of feature engineering
\cite{zheng2013deep,pei2014maxmargin,chen2015gated,chen2015long,cai2016neural,zhang2016transition,yao2016bi,gong2017multi}.
However, it is a waste of resources if we fail to fully exploit all the corpora with different criteria.

\citet{chen2017adversarial} proposed a multi-criteria learning framework for CWS. Specifically, they regard each segmentation criterion as a single task under the framework of multi-task learning~\cite{caruana1997multitask}, where a shared layer is used to extract the criteria-invariant features, and a private layer is used to extract the criteria-specific features. However, it is unnecessary to use a specific private layer for each criterion. These different criterions often have partial overlaps.
As shown in Table \ref{tab:example}, given a sentence ``林丹拿了总冠军 (Lin Dan won the championship)'', the three commonly-used corpora, PKU's People's Daily (PKU) \cite{Yu:2001a}, Penn Chinese Treebank (CTB) \cite{fei2000part} and MSRA \cite{emerson2005second}, use different segmentation criteria. The segmentation of ``林丹 (Lin Dan)'' is the same as in PKU and MSRA criteria, and the segmentation of ``总$|$冠军 (the championship)'' is the same as in CTB and MSRA criteria. All these three criteria have same segmentation for the word ``拿了 (won)''.

\begin{table}\small
\centering
\begin{tabular}{|c|*{5}{c|}}
\hline
Corpora&Lin&Dan&won&\multicolumn{2}{c|}{the championship}\\
\hline
CTB&\multicolumn{2}{c|}{\cellcolor[rgb]{0.8,0.8,0.1}林丹}&拿了&\multicolumn{2}{c|}{总冠军}\\
\hline
PKU&林&丹&拿了&\cellcolor[rgb]{1,0.8,0.8}总&\cellcolor[rgb]{1,0.8,0.8}冠军\\\hline
MSRA&\multicolumn{2}{c|}{\cellcolor[rgb]{0.8,0.8,0.1}林丹}&拿了&\cellcolor[rgb]{1,0.8,0.8}总&\cellcolor[rgb]{1,0.8,0.8}冠军\\
\hline
\end{tabular}
\caption{Illustration of the different segmentation criteria, which partially share some underlying sub-criteria. }\label{tab:example}
\end{table}

Inspired by the above example, we propose a more flexible model for multi-criteria CWS. Each segmentation criterion can be separated into several sub-criteria, each of which is implemented by a long short-term memory neural network (LSTM) \cite{hochreiter1997long}. When segmenting a sentence, these sub-criteria are automatically switched for different words.
Therefore, specific segmentation criteria can be regarded as a combination of these sub-criteria.

Specifically, our model consists of several LSTM cells, each of which represents a segmentation sub-criterion, and a controller, called switcher, to switch the routing among the different LSTMs. These sub-criteria could be shared among different segmentation criteria.

The major merits of our proposed approach are simplicity and flexibility.
Existing work on multi-task learning usually has private layers for each task, parameter, and scale of the computational graph will linearly grow along with the number of tasks. Also, private layers will reduce the flexibility of the model, for example when a new task comes in, approaches with private layers will need to stop training and reconstruct the whole graph. While in our proposed method, all layers can be considered as the shared layer, which will perfectly avoid the growing number of parameter and graph scale. And more interestingly, we can pre-allocate many preserved task embeddings with negligible overhead. With extra task embeddings, we can effortlessly hot-plug a new task into the in-training model.

Finally, we experiment on eight Chinese word segmentation datasets. Experiments show that our models are effective to improve the performance of CWS.

\begin{figure*}[t]
   \centering \includegraphics[width=0.75\textwidth]{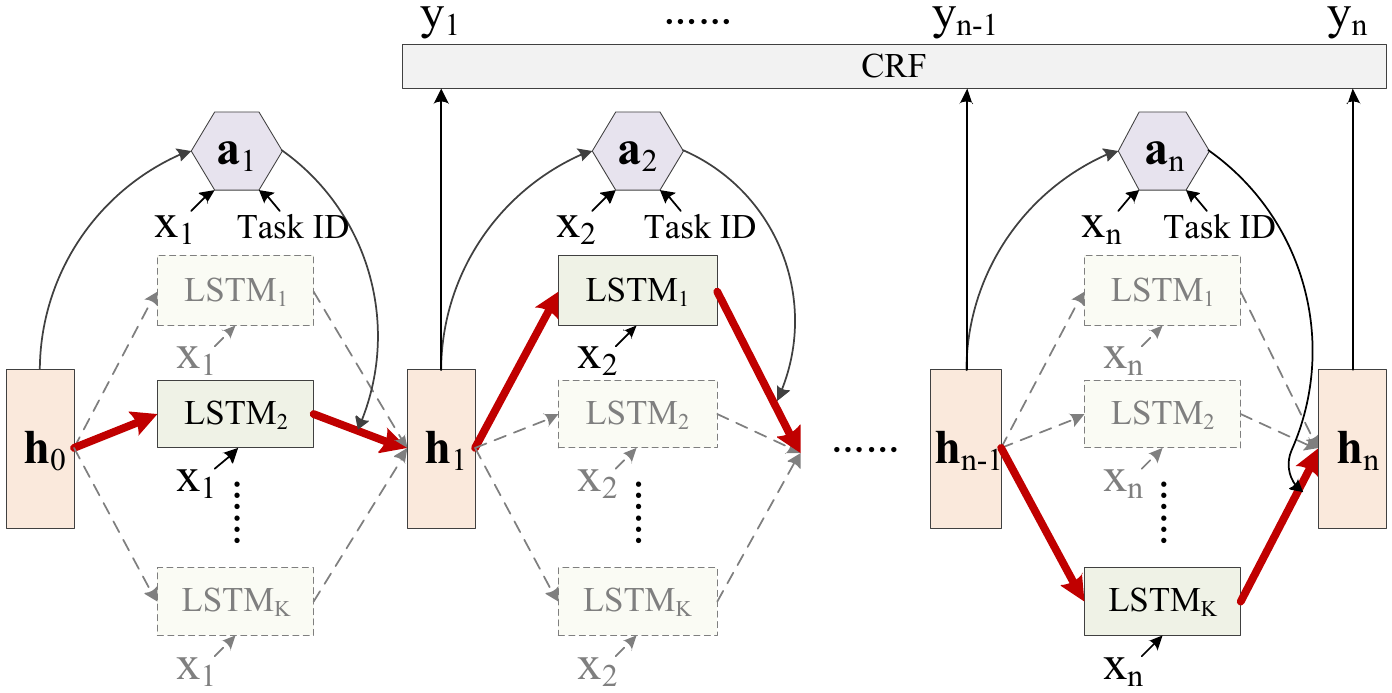}
  \caption{Switch-LSTMs for multi-criteria Chinese word segmentation. $\ba_t$ denote switches taking $\bh_{t-1}$, $x_t$ and task ID $m$ as inputs, and output distributions over k-way LSTMs.} \label{fig:detail_switch_LSTM}
\end{figure*}

\section{Background}
\subsection{Single Criterion CWS with LSTM-CRF Framework}

Chinese word segmentation task could be viewed as a character based sequence labeling problem \cite{zheng2013deep,pei2014maxmargin,chen2015gated,chen2015long}. Specifically, each character in a sentence $X = \{x_1, \dots, x_n\}$ is labelled as one of $\cL =  \{B, M, E, S\}$, indicating the begin, middle, end of a word, or a word with single character. The aim of CWS task is to figure out the ground truth of labels $Y^* = \{y_1^*, \dots, y_n^*\}$:
\begin{equation}
Y^* = \argmax_{Y \in \cL^n} p (Y | X). \label{eq:argmax}
\end{equation}

The popular architecture of neural CWS could be characterized by three components: (1) a character embedding layer; (2) feature extraction layers consisting of several classical neural networks and (3) a CRF tag inference layer. Figure \ref{fig:Model-I} illustrates the general architecture of CWS.

\paragraph{Embedding layer}
In neural models, the first step is to map discrete language symbols into distributed embedding space. Formally, each character $x_t$ is mapped as $\be_{x_t} \in \mathbb{R}^{d_e} \subset \bE$, where $d_e$ is a hyper-parameter indicating the size of character embedding, and $\bE$ is the embedding matrix.

\paragraph{Encoding layers}
Usually a bi-directional LSTM (Bi-LSTMs) \cite{hochreiter1997long} is a prevalent choice for the encoding layer. It could incorporate information from both sides of sequence
\begin{equation}
\bh_t = \BiLSTM(\be_{x_t}, \overrightarrow{\bh}_{t-1}, \overleftarrow{\bh}_{t+1}, \theta),\label{eq:LSTM}
\end{equation}
where $\overrightarrow{\bh}_t$ and $\overleftarrow{\bh}_t$ are the hidden states at step $t$ of the forward and backward LSTMs respectively. $\theta$ denotes all the parameters in Bi-LSTMs layer.

\paragraph{Inference Layer}
The extracted features are then sent to conditional random fields (CRF) \cite{lafferty2001conditional} layer for tag inference. In CRF layer, $p (Y | X)$ in Eq (\ref{eq:argmax}) could be formalized as:
\begin{equation}
p (Y | X) = \frac{\Psi (Y | X)}{\sum_{Y^\prime \in \cL^n} \Psi (Y^\prime | X)}.
\end{equation}
Here, $\Psi (Y | X)$ is the potential function. In first order linear chain CRFs, we have:
\begin{gather}
\Psi (Y | X) = \prod_{t = 2}^n \psi (X, t, y_{t-1}, y_t),\\
\psi (\bx, t, y^\prime, y) = \exp(\delta(X, t)_{y} + \bb_{y^\prime y}),
\end{gather}
where $\bb_{y^\prime y} \in \bR$ is trainable parameters respective to label pair $(y^\prime, y)$. Score function $\delta(X, t) \in \mathbb{R}^{|\cL|}$ calculates scores of each label for tagging the $t$-th character:
\begin{equation}
\delta(X, t) = \bW_{\delta}^\top \mathbf{h}_t + \bb_{\delta}, \label{eq:score}
\end{equation}
where $\mathbf{h}_t$ is the hidden state of Bi-LSTMs at step $t$; $\bW_{\delta} \in \mathbb{R}^{d_h \times |\mathcal{L}|}$ and $\bb_{\delta} \in \mathbb{R}^{|\mathcal{L}|}$ are trainable parameters.

\subsection{Multi-Criteria CWS using Multi-Task Learning}
Annotations in Chinese word segmentation are valuable and expensive, \citet{chen2017adversarial} tried to train on Chinese word segmentation annotations in multiple heterogeneous criteria to improve the performance. The multi-task learning framework is a suitable way to exploit the shared information among these different criteria as shown in Figure \ref{fig:Model-II}.

Formally, assuming that there are $M$ corpora with heterogeneous segmentation criteria, we refer $\mathcal{D}_m$ as corpus $m$ with $N_m$ samples:
\begin{equation}
\mathcal{D}_m = \{(X_i^{(m)},Y_i^{(m)})\}_{i=1}^{N_m},
\end{equation}
where $X_i^m$ and $Y_i^m$ denote the $i$-th sentence and the corresponding label in corpus $m$ respectively.

To exploit information across multiple datasets, the feature layer additionally introduce an shared Bi-LSTMs, together with the original private Bi-LSTMs as in Eq (\ref{eq:LSTM}). Concretely, for corpus $m$, the hidden states of shared layer and private layer are calculated as:
\begin{align}
\bh^{(s)}_t =& \BiLSTM(\be_{x_t}, \overrightarrow{\bh}^{(s)}_{t-1}, \overleftarrow{\bh}^{(s)}_{t+1};\theta^{(s)}_h),\\
\bh^{(m)}_t =& \BiLSTM(\be_{x_t}, \overrightarrow{\bh}^{(m)}_{t-1}, \overleftarrow{\bh}^{(m)}_{t+1};\theta^{(m)}_h).
\end{align}

The score function $\delta(X, t)$ in Eq (\ref{eq:score}) is replaced by:
\begin{equation}
\delta(X, t) = g(\bh^{(s)}_i,\bh^{(m)}_i;\theta^{(m)}_g),
\end{equation}
where $g$ is a feed-forward neural network, taking shared and private features as inputs.

\paragraph{Objective}
The objective is to maximize the log likelihood of true labels on all the corpora:
\begin{equation}
\mathcal{J}_{seg}(\Theta^{m},\Theta^{s}) = \sum_{m=1}^{M} \sum_{i=1}^{N_m}\log p(Y^{(m)}_i|X^{(m)}_i;\Theta^{m},\Theta^{s}),\label{eq:objective}
\end{equation}
where $\Theta^{m}$ = $\{\theta^{(m)}_h,\theta^{(m)}_g\}$ and $\Theta^{s}$ = $\{\bE,\theta^{(s)}_h\}$ denote all the private and shared parameters respectively. 
\begin{table*}[t]\small \setlength{\tabcolsep}{5pt}
\centering
\begin{tabular}{|c|c|c|rrrrrr|}
 \hline
 \multicolumn{3}{|c|}{Corpora} &   \# of Tokens &  \# of Chars &  Dict Size &  Char Types &\# of Sents &  OOV Rate\\\hline
 \hline
  \multirow{4}{*}{\rotatebox{90}{Sighan05}}
  &\multirow{2}*{MSRA}&Train&2.4M& 4.1M& 88.1K& 5.2K& 86.9K&-\\
  &&Test&0.1M& 0.2M& 12.9K& 2.8K& 4.0K&2.60\%\\
  \cline{2-9}
  &\multirow{2}*{AS}&Train&5.4M& 8.4M& 141.3K& 6.1K& 709.0K&-\\
  &&Test&0.1M& 0.2M& 18.8K& 3.7K& 14.4K&4.30\%\\
  \hline
    \hline
\multirow{12}{*}{\rotatebox{90}{Sighan08}}
  &\multirow{2}*{PKU}&Train&  1.1M& 1.8M& 55.2K& 4.7K& 47.3K&-\\
  &&Test&0.2M& 0.3M& 17.6K& 3.4K& 6.4K&3.33\%\\
  \cline{2-9}
  &\multirow{2}*{CTB}&Train&0.6M& 1.1M& 42.2K& 4.2K& 23.4K&-\\
  &&Test&0.1M& 0.1M& 9.8K& 2.6K& 2.1K&5.55\%\\
  \cline{2-9}
  &\multirow{2}*{CKIP}&Train&0.7M& 1.1M& 48.1K& 4.7K& 94.2K&-\\
  &&Test& 0.1M& 0.1M& 15.3K& 3.5K& 10.9K&7.41\%\\
  \cline{2-9}
  &\multirow{2}*{CITYU}&Train&  1.1M& 1.8M& 43.6K& 4.4K& 36.2K&-\\
  &&Test&0.2M& 0.3M& 17.8K& 3.4K& 6.7K&8.23\%\\
  \cline{2-9}
  &\multirow{2}*{NCC}&Train&0.5M& 0.8M& 45.2K& 5.0K& 18.9K&-\\
  &&Test&0.1M& 0.2M& 17.5K& 3.6K& 3.6K&4.74\%\\
  \cline{2-9}
  &\multirow{2}*{SXU}&Train&0.5M& 0.9M& 32.5K& 4.2K& 17.1K&-\\
  &&Test&0.1M& 0.2M& 12.4K& 2.8K& 3.7K&5.12\%\\
  \hline
 \end{tabular}
 \caption{Details of the eight datasets. }\label{tab:info_datasets}
\end{table*}

\section{Switch-LSTMs for Multi-Criteria CWS}
Multi-task learning framework separates its parameters into private ones and shared ones. There are two main drawbacks. First, the model architecture should be manually designed. Second, it is very difficult to generalize to a new tiny dataset. Unlike the multi-task learning framework, switch LSTM doesn't have any private parameters as shown in Figure \ref{fig:Model-III}. Figure \ref{fig:detail_switch_LSTM} gives the architecture of the proposed switch-LSTMs.

\subsection{Switch-LSTMs}
Switch-LSTMs consist of $K$ independent LSTM cells and a switcher. At each time step $t$, switch-LSTMs will switch to one of $K$ LSTMc cells according to the switcher state $\ba_t \in \mathbb{R}^{K}$. Thus, the  switch-LSTMs could be formalized as:
\begin{align}
\bs_{t,k}&=\LSTM(\be_{x_{t}}, \bh_{t-1};\theta^{(s)}_k), \forall k\in[1,K]\\
a_{t,k} &= \Switch(\be_{x_{t}}, \bs_{t,k}, \be_m),\label{eq:switch}\forall k\in[1,K]\nonumber\\
&\triangleq \softmax(\bW [\be_{x_{t}}, \bs_{t,k}, \be_m]),\\
\bh_{t} 
&= \sum_{k=1}^{K} a_{t,k} \bs_{t,k},
\end{align}
where $\theta^{(s)} = \{\theta^{(s)}_1,\theta^{(s)}_2,\cdots, \theta^{(s)}_K\}$ denotes the parameter of the corresponding LSTM. Since the switcher picks layers according to the task property as well, we introduce a task embedding $\be_m \subset \bE^m$ for task ID $m$.

We abbreviate the above equations of switch-LSTMs to
\begin{align}
\bh_{t}&=\SwitchLSTM(\be_{x_{t}}, \bh_{t-1};\theta^{(s)}).\label{eq:adaLSTM}
\end{align}

Similar to Bi-LSTMs, we use bi-directional switch LSTMs (Bi-Switch-LSTMs) for multi-criteria Chinese word segmentation. The feature $\bh_t$ extracted by Bi-Switch-LSTMs could be formalized as:
\begin{align}
\bh_t &= \BiSwitchLSTM(\be_{x_i}, \overrightarrow{\bh}_{t-1}, \overleftarrow{\bh}_{t+1}),
\end{align}
where $\overrightarrow{\bh}_{t}$ and $\overleftarrow{\bh}_{t}$ are forward  and backward adaptive LSTMs respectively as in Eq (\ref{eq:adaLSTM}). Thus, the forward and backward switch gate statuses are $\overrightarrow{A} = \{\overrightarrow{\ba}_1,\cdots,\overrightarrow{\ba}_n\}$ and
$\overleftarrow{A} = \{\overleftarrow{\ba}_1,\cdots,\overleftarrow{\ba}_n\}$ respectively.

The score function $\delta(X, t)$ in Eq (\ref{eq:score}) is replaced by:
\begin{equation}
\delta(X, t)= \sum_{i=1}^K\overrightarrow{\ba}_{t,i}\overrightarrow{\bW}_{i}\overrightarrow{\bh}_{t} + \sum_{i=1}^K\overleftarrow{\ba}_{t,i}\overleftarrow{\bW}_{i}\overleftarrow{\bh}_{t},
\end{equation}

where $+$ denotes the element-wise addition operation,
$\overleftarrow{\bW}_{i}$ and $\overrightarrow{\bW}_{i}$ are parameters for $i$th LSTM cells in \textit{left} and \textit{right} Switch-LSTMs respectively.

\subsection{Objective for Multi-task Learning}
The objective is to maximize the log likelihood $\cJ_{seg}(\Theta)$:
\begin{equation}
\mathcal{J}_{seg}(\Theta) = \sum_{m=1}^{M} \sum_{i=1}^{N_m}\log p(Y^{(m)}_i|X^{(m)}_i,m;\Theta),
\end{equation}
where $\Theta = \{\bE, \theta^{(s)}, \bE^m, \bW, \overrightarrow{\bW}, \overleftarrow{\bW}\}$ are all shared trainable parameters.

\section{Experiments}

\subsection{Datasets}
  We experiment on eight CWS datasets from SIGHAN2005 \cite{emerson2005second} and SIGHAN2008 \cite{moe2008fourth}. Table \ref{tab:info_datasets} gives the details of the eight datasets. AS, CITYU and CKIP are in traditional Chinese. MSRA, PKU, CTB, NCC and SXU are in simplified Chinese. We randomly pick 10\% instances from training set as the development set for all the datasets.

\subsection{Experimental Configuration}
    The character embedding size $d_e$ is set to 100, task embedding size is set to 20, the hidden size $d_h$ for our proposed Switch-LSTMs are set to 100, the number of choices $K$ in K-way switch is set to one of \{1, 2, 3, 4, 5, 6, 7, 8, 9, 10\}.  As a common approach to alleviate overfitting, we dropout our embedding with a probability of 0.2. Other than embedding, we use Xavier uniform initializer for all trainable parameters in our model. All traditional Chinese characters are mapped to the simplified Chinese character, and all experiments are conducted upon pre-trained character embedding with bigram feature, embeddings are pre-trained with word2vec toolkit \cite{Mikolov:2013} on Chinese Wikipedia corpus. Character embeddings are shared across tasks.

\subsection{Training}
    For each training step, we sample 6 tasks from the task pool, each with a batch size of 128, the probability of being sampled is proportional to the total number of samples in the corresponding task. Then we feed 6 batches of data into 6 computational graphs and then update the global parameter synchronously. For parameter selection, we keep track of parameters that perform the best on the development set. The training process is terminated after 7 epochs without improvement on development set, and note that we use an averaged $F$ scores over all tasks to measure how good our model is.

\subsection{Overall Results}
\begin{table*}[t]\small 
\centering
\begin{tabular}{|c|*{9}{c|}>{\columncolor[gray]{.8}}c|}
\hline
  \multicolumn{2}{|c|}{Models}&
MSRA  &AS &PKU  &CTB  &CKIP &CITYU  &NCC  &SXU  &Avg.\\
\hline
\multicolumn{11}{|l|}{Single-Criterion Learning } \\
\hline
\multirow{4}*{LSTM} &P   &95.13     &93.66     &93.96     &95.36     &91.85     &94.01     &91.45     &95.02     &93.81\\
 &R   &95.55     &94.71     &92.65     &85.52     &93.34     &94.00     &92.22     &95.05     &92.88\\
 \cdashline{2-11}
 &F   &95.34     &94.18     &93.30     &95.44     &92.59     &94.00     &91.83     &95.04     &93.97\\ \cdashline{2-11}
 &OOV   &63.60     &69.83     &66.34     &76.34     &68.67     &65.48     &56.28     &69.46     &67.00 \\
 \hline
\multirow{4}*{\tabincell{c}{Bi-LSTMs\\\cite{chen2017adversarial}}} &P  &95.70  &93.64  &93.67  &95.19  &92.44  &94.00  &91.86  &95.11  &93.95  \\
&R  &95.99  &94.77  &92.93  &95.42  &93.69  &94.15  &92.47  &95.23  &94.33  \\
\cdashline{2-11}
&F  &95.84  &94.20  &93.30  &95.30  &93.06  &94.07  &92.17  &95.17  &94.14  \\\cdashline{2-11}
&OOV  &66.28  &70.07  &66.09  &76.47  &72.12  &65.79  &59.11  &71.27  &68.40  \\
\hline
\multirow{4}*{\tabincell{c}{Stacked Bi-LSTM\\\cite{chen2017adversarial}}} &P  &95.69     &93.89     &94.10     &95.20     &92.40     &94.13     &91.81     &94.99     &94.03\\
&R  &95.81     &94.54     &92.66     &95.40     &93.39     &93.99     &92.62     &95.37     &94.22\\
\cdashline{2-11}
&F  &95.75     &94.22     &93.37     &95.30     &92.89     &94.06     &92.21     &95.18     &94.12\\
\cdashline{2-11}
&OOV  &65.55     &71.50     &67.92     &75.44     &70.50     &66.35     &57.39     &69.69     &68.04\\
\hline
\multirow{4}*{\tabincell{c}{Switch-LSTMs\\This Work}}
&P&96.07&93.83&95.92&97.13&92.02&93.69&91.81&95.02&94.44\\
&R&96.86&95.21&95.56&97.05&93.76&93.73&92.43&96.13&95.09\\
\cdashline{2-11}
&F&96.46&94.51&95.74&97.09&92.88&93.71&92.12&95.57&94.76\\
\cdashline{2-11}
&OOV&69.90&77.80&72.70&81.80&71.60&59.80&55.50&67.30&69.55\\
\hline
\multicolumn{11}{|l|}{Multi-Criteria Learning } \\
\hline
\multirow{4}*{Bi-LSTMs}
&P&94.64&93.54&93.24&92.87&93.26&91.41&89.30&92.61&92.61\\
&R&93.20&94.06&91.94&91.75&93.41&90.64&88.04&92.42&91.93\\
\cdashline{2-11}
&F&93.91&93.80&92.59&92.31&93.33&91.02&88.66&92.51&92.27\\
\cdashline{2-11}
&OOV&65.60&89.20&64.90&75.40&80.00&74.80&64.00&68.50&72.80\\
\hline
\multirow{4}*{\tabincell{c}{Multi-Task Framework\\\cite{chen2017adversarial}} } &P  &95.76  &93.99  &94.95  &95.85  &93.50  &95.56  &92.17  &96.10  &94.74  \\
&R  &95.89  &95.07  &93.48  &96.11  &94.58  &95.62  &92.96  &96.13  &94.98  \\
\cdashline{2-11}
&F  &95.82  &94.53  &94.21  &95.98  &94.04  &95.59  &92.57  &96.12  &94.86  \\
\cdashline{2-11}
&OOV  &70.72  &72.59  &73.12  &81.21  &76.56  &82.14  &60.83  &77.56  &74.34  \\
\hline
\multirow{4}*{\tabincell{c}{Switch-LSTMs\\This Work}}
&P&97.69&94.42&96.24&97.09&94.53&95.85&94.07&96.88&95.85\\
&R&97.87&96.03&96.05&97.43&95.45&96.59&94.17&97.62&96.40\\
\cdashline{2-11}
&F&\textbf{97.78}&\textbf{95.22}&\textbf{96.15}&\textbf{97.26}&\textbf{94.99}&\textbf{96.22}&\textbf{94.12}&\textbf{97.25}&\textbf{96.12}\\
\cdashline{2-11}
&OOV&64.20&77.33&69.88&83.89&77.69&73.58&69.76&78.69&74.38\\
\hline
\end{tabular}
\caption{Results of the proposed model on the test sets of eight CWS datasets.
There are two blocks. The first block consists of single-criterion learning models. LSTM and Bi-LSTMs are baselines and the results on them are reported in \citet{chen2017adversarial}. The second block consists of the multi-criteria learning model. Multi-task framework for multi-criterion Chinese word segmentation is proposed by \citet{chen2017adversarial}. Here, P, R, F, OOV indicate the precision, recall, F value and OOV recall rate respectively. The maximum F values are highlighted for each dataset.
}\label{tab:res}
\end{table*}
Switch-LSTMs could work on both single-criterion learning and multi-criteria learning scenarios. Table \ref{tab:res} gives the overall results of the proposed model. Here, Switch-LSTMs have 4-way switches. When applying Switch-LSTMs to single-criterion learning scenario, we omit the $\be_m$ term in Eq (\ref{eq:switch}) and the action $\ba_t$ is only related to the current state $\bs_t$ and the input $\be_{x_{t+1}}$.

In \textit{single-criterion learning} scenario, we compare Switch-LSTMs with prevalent LSTM and Bi-LSTMs model. As we can see, Bi-LSTMs outperform LSTM and Switch-LSTMs outperform previous LSTM and Bi-LSTMs models. Quantitatively speaking, Switch-LSTMs obtain 94.76 in average F-value, while LSTM and Bi-LSTMs obtain 93.97 and 94.14 in average F-value respectively. It shows that the capability of Switch-LSTMs is greater than conventional LSTM.

In \textit{multi-criteria learning} scenario, we compare Switch-LSTMs with Bi-LSTMs and multi-task learning framework (MTL) proposed by \citet{chen2017adversarial}. Notably, Bi-LSTMs is a special case of Switch-LSTM with $K$ set to 1 (single-criterion learning). By concatenating all datasets, Bi-LSTMs performs poorly in multi-criteria learning scenario (the worst). Experimental results show that Switch-LSTMs outperform both Bi-LSTMs and multi-task learning framework on all the corpora. In average, Switch-LSTMs boost about +1\% (96.12 in F-value) compared to  multi-task learning framework (94.86 in F-value), and boosts +3.85\% compared to Bi-LSTMs model (92.27 in F-value).

We could also observe that the performance benefits from multi-criteria learning, since, in this case, the model could learn extra helpful information from other corpora. Concretely, in average F-value, Switch-LSTMs for multi-criteria learning boosts +1.36\% (96.12 in F-value) compared to Switch-LSTMs for single-criterion learning (94.76 in F-value).

\subsection{Model Selection}\label{sec:model_selection}

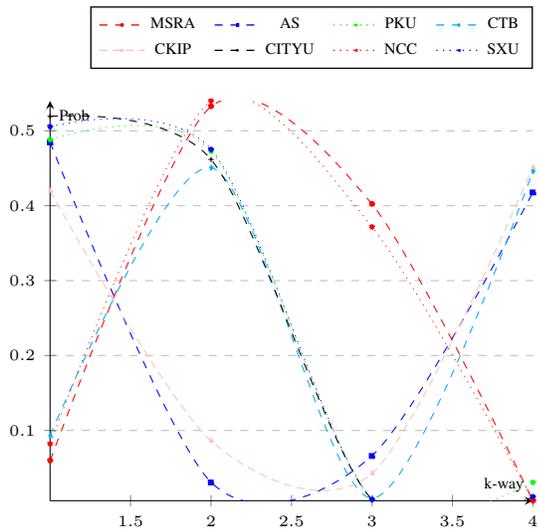
\begin{figure}[b!]
  \centering
  \pgfplotsset{width=0.45\textwidth}
  \begin{tikzpicture}
    \begin{axis}[
    xlabel={k-way},
    ylabel={Prob},
    legend entries={MSRA,    AS,    PKU,    CTB,    CKIP,    CITYU,    NCC,    SXU, Avg},
    mark size=1.0pt,
    ymajorgrids=true,
    grid style=dashed,
    legend pos= south east,
    legend style={font=\tiny,line width=.5pt,mark size=.5pt,
            at={(1,1.08)},
            legend columns=4,
            /tikz/every even column/.append style={column sep=0.5em}},
            smooth,
    ]
    \addplot [red,dashed,mark=*] table [x index=0, y index=1] {axis/switch_forward.txt};
    \addplot [blue,dashed,mark=square*] table [x index=0, y index=2] {axis/switch_forward.txt};
    \addplot [green,dotted,mark=otimes*] table [x index=0, y index=3] {axis/switch_forward.txt};
    \addplot [cyan,dashed,mark=diamond*] table [x index=0, y index=4] {axis/switch_forward.txt};
    \addplot [pink,densely dashed,mark=triangle*] table [x index=0, y index=5] {axis/switch_forward.txt};
    \addplot [black,dashed,,mark=+] table [x index=0, y index=6] {axis/switch_forward.txt};
    \addplot [red,dotted,mark=*] table [x index=0, y index=7] {axis/switch_forward.txt};
    \addplot [blue,dotted,mark=*] table [x index=0, y index=8] {axis/switch_forward.txt};
    \end{axis}
\end{tikzpicture}
\caption{Distributions of switch statuses of forward Switch-LSTMs on test sets of eight datasets for multi-criteria learning. X-axis denotes every individual switch state, and Y-axis is the corresponding probability that state occurs.}\label{fig:visualization}
\end{figure}

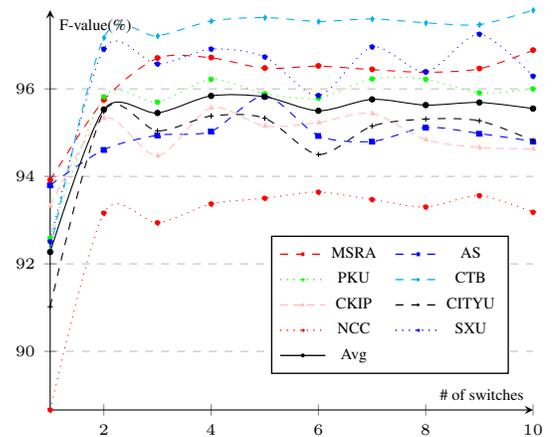
\begin{figure}[t]
  \centering
  \pgfplotsset{width=0.45\textwidth}
  \begin{tikzpicture}
    \begin{axis}[
    xlabel={\# of switches},
    ylabel={F-value(\%)},
    legend entries={MSRA,    AS,    PKU,    CTB,    CKIP,    CITYU,    NCC,    SXU, Avg},
    mark size=1.0pt,
    ymajorgrids=true,
    grid style=dashed,
    legend pos= south east,
    legend style={font=\tiny,line width=.5pt,mark size=.5pt,
            at={(0.95,0.09)},
            legend columns=2,
            /tikz/every even column/.append style={column sep=0.5em}},
            smooth,
    ]
    \addplot [red,dashed,mark=*] table [x index=0, y index=1] {axis/multi_no_switch_f.txt};
    \addplot [blue,dashed,mark=square*] table [x index=0, y index=2] {axis/multi_no_switch_f.txt};
    \addplot [green,dotted,mark=otimes*] table [x index=0, y index=3] {axis/multi_no_switch_f.txt};
    \addplot [cyan,dashed,mark=diamond*] table [x index=0, y index=4] {axis/multi_no_switch_f.txt};
    \addplot [pink,densely dashed,mark=triangle*] table [x index=0, y index=5] {axis/multi_no_switch_f.txt};
    \addplot [black,dashed,,mark=+] table [x index=0, y index=6] {axis/multi_no_switch_f.txt};
    \addplot [red,dotted,mark=*] table [x index=0, y index=7] {axis/multi_no_switch_f.txt};
    \addplot [blue,dotted,mark=*] table [x index=0, y index=8] {axis/multi_no_switch_f.txt};
    \addplot [black,solid,mark=*] table [x index=0, y index=9] {axis/multi_no_switch_f.txt};
    \end{axis}
\end{tikzpicture}
\caption{Results (on F-value) of Switch-LSTMs with various numbers of switches on development sets of eight datasets on multi-criteria learning scenario. X-axis denotes the \# of switches that the model employs, and Y-axis reports the corresponding results.}\label{fig:dev_switch}
\end{figure}

Figure \ref{fig:dev_switch} shows the relationship between switch number and performance in the multi-criteria learning scenario. As we can see, models with more than 2 switches are better than 1-switch-LSTM with a considerable margin, and the case with 4-way switches is slightly better than other settings. So we employ 4-way Switch-LSTMs for the following experiments. 1-way Switch-LSTMs are the traditional LSTM. So, LSTM could be viewed as a special case of the proposed Switch-LSTMs.

\subsection{Scale of Parameter Set}
\begin{table*}[t] \setlength{\tabcolsep}{5.5pt}
\centering
\begin{tabular}{|c|*{10}{c|}>{\columncolor[gray]{.8}}c|}
\hline
  \multicolumn{2}{|c|}{Models}&\# of Param.&
MSRA  &AS &PKU  &CTB  &CKIP &CITYU  &NCC  &SXU  &Avg.\\
\hline
 \multicolumn{2}{|c|}{Multi-Task Framework}&25K
  &95.82  &94.53  &94.21  &95.98  &94.04  &95.59  &92.57  &96.12  &94.86  \\
\hline
\multirow{10}*{\tabincell{c}{Switch-LSTMs\\\# of switches}}
&1&4K&93.91&93.80&92.59&92.31&93.33&91.02&88.66&92.51&92.27\\
&2&7K&95.74&94.61&95.82&97.17&95.33&95.48&93.16&96.91&95.53\\
&3&11K&96.70&94.94&95.70&97.21&94.47&95.04&92.94&96.57&95.45\\
&4&15K&96.71&95.03&96.22&97.55&95.57&95.38&93.37&96.91&\textbf{95.84}\\
&5&18K&96.47&95.84&95.89&97.63&95.15&95.33&93.50&96.73&95.82\\
&6&22K&96.52&94.93&95.79&97.54&95.23&94.50&93.64&95.85&95.50\\
&7&25K&96.44&94.80&96.23&97.60&95.44&95.15&93.47&96.96&95.76\\
&8&29K&96.38&95.12&96.22&97.51&94.84&95.31&93.30&96.39&95.63\\
&9&33K&96.46&94.98&95.91&97.47&94.66&95.27&93.56&97.25&95.69\\
&10&36K&96.88&94.80&96.00&97.80&94.63&94.82&93.18&96.29&95.55\\
\hline
\end{tabular}
\caption{Results of multi-task framework and Switch-LSTMs with various numbers of switches on the test sets of eight datasets on multi-criteria learning scenario. Only F-values are reported. \# of parameters only takes parameters of the model into account (input embedding matrix $\bE$ is not included).}\label{tab:test_switch}
\end{table*}
Table \ref{tab:test_switch} gives the results of multi-task framework and Switch-LSTMs on the test sets of eight datasets for multi-criteria learning. For 8 datasets, the multi-task framework contains 8 private Bi-LSTMs and 1 shared Bi-LSTMs, whereas Switch-LSTMs do not have any private parameters, consisting of $K$ LSTM cells associated with one switch for control. As we could observe, the parameter set size of multi-task framework is 25K, while the parameter set size of Switch-LSTMs ranges from 4K to 36K with respect to various number of switches. However, as mentioned,
Switch-LSTMs perform great when we have more than 2 switches. Concretely, 2-way Switch-LSTMs obtain 95.53 on F-value averagely, outperforming the multi-task framework (94.86 on F-value). But the parameter set size of 2-way Switch-LSTMs is only 7K. Therefore, Switch-LSTMs could outperform multi-task framework with fewer parameters.

\subsection{Visualization}



To better understand how switches affect and contribute to the performance, we visualize the statuses of switches for eight datasets as shown in Figure \ref{fig:visualization}. Concretely, each switch gives a distribution over k-way LSTM on all step $t$ for all the instances. Thus, we calculate the global switch distribution by a normalization operation for each dataset. As we observe, switch distributions are diverse, and datasets with similar criteria obtain similar switch distributions. Due to space limitation, we only plot the switch statuses of forwarding Switch-LSTMs.

\subsection{Effects of Switch}
\begin{table*}[t] 
\centering
\begin{tabular}{|l|*{8}{c}>{\columncolor[gray]{.8}}c|}
\hline
Models&
MSRA  &AS &PKU  &CTB  &CKIP &CITYU  &NCC  &SXU  &Avg.\\
\hline
w/o states $\bs_t$&95.68&95.03&95.71&\textbf{97.60}&94.73&94.84&93.34&96.58&95.44\\
w/o input embeddings $\be_{x_i}$&96.28&94.82&95.98&97.18&94.91&95.14&\textbf{93.46}&96.54&95.54\\
w/o task embedding $\be_m$&93.36&93.49&92.90&92.65&93.81&91.61&89.25&93.02&92.51\\
w/o $\be_{x_i}$ \& $\be_m$&93.26&93.50&92.60&93.31&94.23&91.74&88.74&93.10&92.56\\
w/o $\bs_t$ \& $\be_m$&93.39&91.97&91.66&92.51&92.09&91.22&88.43&92.33&91.70\\
\hline
Randomly switch in training&92.64&92.90&92.84&93.51&94.20&91.63&89.45&93.05&92.53\\
Randomly switch in testing&85.56&93.63&90.16&97.16&93.53&93.22&85.56&91.49&91.29\\
\hline
All (Full model)&\textbf{96.71}&\textbf{95.03}&\textbf{96.22}&97.55&\textbf{95.57}&\textbf{95.38}&93.37&\textbf{96.91}&\textbf{95.84}\\
\hline
\end{tabular}

\caption{Results of the proposed model with different switch configurations on test sets of eight CWS datasets for multi-criteria learning. Only F-values are reported. Term ``w/o'' denotes ``without''. The maximum F values are highlighted for each dataset.
}\label{tab:effect_switch}
\end{table*}

\begin{table*}[t] 
\centering
\begin{tabular}{|c|*{9}{c|}>{\columncolor[gray]{.8}}c|}
\hline
  \# of instances & Models&
MSRA  &AS &PKU  &CTB  &CKIP &CITYU  &NCC  &SXU  &Avg.\\
\hline
\multirow{2}*{100}
&Single&75.71&68.51&78.87&79.65&65.59&77.18&71.75&81.29&74.82\\
&Transfer&89.08&91.93&91.50&92.08&94.02&93.62&88.97&93.34&91.82\\
\hline
\multirow{2}*{300}
&Single&81.50&72.77&83.95&88.46&73.76&82.12&78.20&85.97&80.84\\
&Transfer&88.99&92.05&91.97&91.92&94.52&93.43&89.94&94.03&92.11\\
\hline
\multirow{2}*{500}
&Single&83.52&75.29&86.02&90.33&75.12&86.43&79.89&87.12&82.97\\
&Transfer&89.17&92.23&91.90&94.94&94.61&93.72&89.98&\textbf{94.19}&92.59\\
\hline
\multirow{2}*{700}
&Single&83.55&76.29&87.70&91.59&77.76&87.97&81.72&88.33&75.25\\
&Transfer&\textbf{89.41}&92.10&91.74&\textbf{95.49}&94.49&\textbf{93.87}&90.12&93.67&92.61\\
\hline
\multirow{2}*{1000}
&Single&85.75&78.78&88.75&91.82&78.51&88.70&82.52&90.09&76.30\\
&Transfer&89.04&\textbf{92.35}&\textbf{92.51}&95.39&\textbf{94.64}&93.82&\textbf{90.45}&93.85&\textbf{92.76}\\
\hline
\end{tabular}
\caption{Results of transfer learning of the proposed model on test sets of eight CWS datasets for single-criteria learning. Only F-values are reported. ``\# of instances'' denotes how many instances involved for training. Single model is the conventional Bi-LSTM model. ``Transfer'' denotes the proposed Switch-LSTMs by fixing all the parameters learned from other 7 datasets except the new involved task embedding.
}\label{tab:transfer}
\end{table*}

To evaluate how the inputs of switch affect the performance, we do some input ablation experiments. As shown in Table \ref{tab:effect_switch}, all the inputs of switch contribute to the model performance, and the task embedding $\be_m$ is the most effective part. Concretely, without task embedding $\be_m$, the average F-value over eight datasets drops by 3.33\% compared to the full model (95.84 on F-value). Besides, states $\bs_t$ are also crucial to switch. The performance further drops by 0.81\% in average F-value (from 92.51 to 91.70) by additionally ablating state information $\bs_t$.

We also tried two switch strategies to see how the switch affect the performance: randomly switch in the training phase and randomly switch in the testing phase. Former one always randomly picks up a switch status at each time step during training, and when at test phase, a normal switch is employed. The later one train the switch in a normal way but randomly picks up a switch status at each time step at the testing phase. By randomly pick up a switch at each time, we could test that if the model works well without the switch. As we observe, random strategies on switch lead to poor performance (performance drops by 3\% on F-value). It implies that a normal switch is crucial to our model, and switch mechanism contributes a lot in boosting performance.






\subsection{Knowledge Transfer}


Switch-LSTMs could also be easily transferred to other new datasets. To evaluate the transfer capability of our model, we leave one dataset out and train Switch-LSTMs on other 7 datasets. Then, we fixed all the parameters except the newly introduced task embedding when training on instances of the leave out dataset. As shown in Table \ref{tab:transfer}, Switch-LSTMs could obtain excellent performance when only 100, 300. 500, 700, 1000 training instances are available. The single model (conventional LSTM) cannot learn from such few instances. For instance, when we train with 1000 training examples, the single model only obtains 76.30 in average F-value, whereas Switch-LSTMs could obtain 92.76 (boosts 16.46 averagely). It shows that Switch-LSTMs could adapt to a new criterion by only learning a new task embedding, and the newly learned task embedding leads to a new switch strategy for the new criterion.

\section{Related Work}
It is a common practice for utilizing annotated data from different but similar domain to boost the performance for each task. Many efforts have been made to better utilizing the homogeneous factor in various tasks to help improve multiple tasks especially those barren tasks with few examples.

Recently, some efforts have been made to transfer knowledge between NLP Tasks. \citet{zoph2016multisourcemt,Melvin2017googleMultiMt} have been jointly training translation models from and to different languages, it is achieved simply by jointly train encoder or both encoder and decoder. \cite{Jiang:2009,sun2012reducing,qiu2013joint,li-EtAl:2015:ACL-IJCNLP3,li2016fast,chenneural} adopted the stack-based model to take advantage of annotated data from multiple sources, and show that tasks can indeed help improve each other.

\citet{chenneural} adopted two neural models based on stacking framework and multi-view framework respectively, which boosts POS-tagging performance by utilizing corpora in heterogeneous annotations.
\citet{chen2017adversarial} have proposed a multi-criteria learning framework for CWS. Using a similar framework as in \citet{caruana1997multitask}, there are private layers for each task to extract criteria-specific features, and a shared layer for the purpose of transferring information between tasks, to avoid negative transfer, they pose an adversarial loss on the shared layer to impose source indistinguishability thus make it criteria-invariant.

\section{Conclusions}

In this paper, we propose a flexible model, called Switch-LSTMs, for multi-criteria CWS, which can improve the performance of every single criterion by fully exploiting the underlying shared sub-criteria across multiple heterogeneous criteria. Experiments on eight corpora show the effectiveness of our proposed model.
In future works, we are planning to experiment Switch-LSTMs on other multi-task problems such as transferring information between different document categorization datasets and further investigate a discrete version of Switch-LSTM via reinforcement learning.

\section*{Acknowledgments}
We would like to thank the anonymous reviewers for their valuable comments. The research work is supported by Shanghai Municipal Science and Technology Commission (No. 17JC1404100 and 16JC1420401), and National Natural Science Foundation of China (No. 61672162 and 61751201).


\end{CJK*}

\bibliography{nlp}

\begin{thebibliography}{}

\bibitem[\protect\citeauthoryear{Cai and Zhao}{2016}]{cai2016neural}
Cai, D., and Zhao, H.
\newblock 2016.
\newblock Neural word segmentation learning for chinese.
\newblock In {\em Proceedings of Annual Meeting of the Association for
  Computational Linguistics, ACL}.

\bibitem[\protect\citeauthoryear{Caruana}{1997}]{caruana1997multitask}
Caruana, R.
\newblock 1997.
\newblock Multitask learning.
\newblock {\em Machine learning}.

\bibitem[\protect\citeauthoryear{Chen \bgroup et al\mbox.\egroup
  }{2015a}]{chen2015gated}
Chen, X.; Qiu, X.; Zhu, C.; and Huang, X.
\newblock 2015a.
\newblock Gated recursive neural network for chinese word segmentation.
\newblock In {\em Proceedings of Annual Meeting of the Association for
  Computational Linguistics, ACL}.

\bibitem[\protect\citeauthoryear{Chen \bgroup et al\mbox.\egroup
  }{2015b}]{chen2015long}
Chen, X.; Qiu, X.; Zhu, C.; Liu, P.; and Huang, X.
\newblock 2015b.
\newblock Long short-term memory neural networks for chinese word segmentation.
\newblock In {\em Proceedings of the Conference on Empirical Methods in Natural
  Language Processing, EMNLP}.

\bibitem[\protect\citeauthoryear{Chen \bgroup et al\mbox.\egroup
  }{2017}]{chen2017adversarial}
Chen, X.; Shi, Z.; Qiu, X.; and Huang, X.
\newblock 2017.
\newblock Adversarial multi-criteria learning for chinese word segmentation.
\newblock In {\em Proceedings of Annual Meeting of the Association for
  Computational Linguistics, ACL}.

\bibitem[\protect\citeauthoryear{Chen, Zhang, and Liu}{2016}]{chenneural}
Chen, H.; Zhang, Y.; and Liu, Q.
\newblock 2016.
\newblock Neural network for heterogeneous annotations.
\newblock {\em Proceedings of the Conference on Empirical Methods in Natural
  Language Processing, EMNLP}.

\bibitem[\protect\citeauthoryear{Emerson}{2005}]{emerson2005second}
Emerson, T.
\newblock 2005.
\newblock The second international chinese word segmentation bakeoff.
\newblock In {\em Proceedings of the fourth SIGHAN workshop on Chinese language
  Processing}.

\bibitem[\protect\citeauthoryear{Fei}{2000}]{fei2000part}
Fei, X.
\newblock 2000.
\newblock The part-of-speech tagging guidelines for the penn chinese treebank
  (3.0).
\newblock {\em URL: http://www. cis. upenn. edu/\~{} chinese/segguide. 3rd. ch.
  pdf}.

\bibitem[\protect\citeauthoryear{Gong \bgroup et al\mbox.\egroup
  }{2017}]{gong2017multi}
Gong, C.; Li, Z.; Zhang, M.; and Jiang, X.
\newblock 2017.
\newblock Multi-grained chinese word segmentation.
\newblock In {\em Proceedings of the Conference on Empirical Methods in Natural
  Language Processing, EMNLP}.

\bibitem[\protect\citeauthoryear{Hochreiter and
  Schmidhuber}{1997}]{hochreiter1997long}
Hochreiter, S., and Schmidhuber, J.
\newblock 1997.
\newblock Long short-term memory.
\newblock {\em Neural computation}.

\bibitem[\protect\citeauthoryear{Jiang, Huang, and Liu}{2009}]{Jiang:2009}
Jiang, W.; Huang, L.; and Liu, Q.
\newblock 2009.
\newblock Automatic adaptation of annotation standards: Chinese word
  segmentation and {POS} tagging: a case study.
\newblock In {\em Proceedings of the Joint Conference of the 47th Annual
  Meeting of the ACL and the 4th International Joint Conference on Natural
  Language Processing}.

\bibitem[\protect\citeauthoryear{Jin and Chen}{2008}]{moe2008fourth}
Jin, G., and Chen, X.
\newblock 2008.
\newblock The fourth international chinese language processing bakeoff: Chinese
  word segmentation, named entity recognition and chinese pos tagging.
\newblock In {\em Sixth SIGHAN Workshop on Chinese Language Processing}.

\bibitem[\protect\citeauthoryear{Johnson \bgroup et al\mbox.\egroup
  }{2017}]{Melvin2017googleMultiMt}
Johnson, M.; Schuster, M.; Le, Q.~V.; Krikun, M.; Wu, Y.; Chen, Z.; Thorat, N.;
  Vi{\'{e}}gas, F.~B.; Wattenberg, M.; Corrado, G.; Hughes, M.; and Dean, J.
\newblock 2017.
\newblock Google's multilingual neural machine translation system: Enabling
  zero-shot translation.
\newblock {\em {TACL}}.

\bibitem[\protect\citeauthoryear{Lafferty, McCallum, and
  Pereira}{2001}]{lafferty2001conditional}
Lafferty, J.~D.; McCallum, A.; and Pereira, F. C.~N.
\newblock 2001.
\newblock Conditional random fields: Probabilistic models for segmenting and
  labeling sequence data.
\newblock In {\em Proceedings of the Eighteenth International Conference on
  Machine Learning}.

\bibitem[\protect\citeauthoryear{Li \bgroup et al\mbox.\egroup
  }{2015}]{li-EtAl:2015:ACL-IJCNLP3}
Li, Z.; Chao, J.; Zhang, M.; and Chen, W.
\newblock 2015.
\newblock Coupled sequence labeling on heterogeneous annotations: Pos tagging
  as a case study.
\newblock In {\em Proceedings of the 53rd Annual Meeting of the Association for
  Computational Linguistics and the 7th International Joint Conference on
  Natural Language Processing}.

\bibitem[\protect\citeauthoryear{Li \bgroup et al\mbox.\egroup
  }{2016}]{li2016fast}
Li, Z.; Chao, J.; Zhang, M.; and Yang, J.
\newblock 2016.
\newblock Fast coupled sequence labeling on heterogeneous annotations via
  context-aware pruning.
\newblock In {\em Proceedings of the Conference on Empirical Methods in Natural
  Language Processing, EMNLP}.

\bibitem[\protect\citeauthoryear{Mikolov \bgroup et al\mbox.\egroup
  }{2013}]{Mikolov:2013}
Mikolov, T.; Sutskever, I.; Chen, K.; Corrado, G.~S.; and Dean, J.
\newblock 2013.
\newblock Distributed representations of words and phrases and their
  compositionality.
\newblock In {\em Advances in Neural Information Processing Systems}.

\bibitem[\protect\citeauthoryear{Pei, Ge, and Baobao}{2014}]{pei2014maxmargin}
Pei, W.; Ge, T.; and Baobao, C.
\newblock 2014.
\newblock Maxmargin tensor neural network for chinese word segmentation.
\newblock In {\em Proceedings of Annual Meeting of the Association for
  Computational Linguistics, ACL}.

\bibitem[\protect\citeauthoryear{Qiu, Zhao, and Huang}{2013}]{qiu2013joint}
Qiu, X.; Zhao, J.; and Huang, X.
\newblock 2013.
\newblock Joint chinese word segmentation and pos tagging on heterogeneous
  annotated corpora with multiple task learning.
\newblock In {\em Proceedings of the Conference on Empirical Methods in Natural
  Language Processing, EMNLP}.

\bibitem[\protect\citeauthoryear{Sun and Wan}{2012}]{sun2012reducing}
Sun, W., and Wan, X.
\newblock 2012.
\newblock Reducing approximation and estimation errors for chinese lexical
  processing with heterogeneous annotations.
\newblock In {\em Proceedings of Annual Meeting of the Association for
  Computational Linguistics, ACL}.

\bibitem[\protect\citeauthoryear{Yao and Huang}{2016}]{yao2016bi}
Yao, Y., and Huang, Z.
\newblock 2016.
\newblock Bi-directional lstm recurrent neural network for chinese word
  segmentation.
\newblock In {\em International Conference on Neural Information Processing}.

\bibitem[\protect\citeauthoryear{Yu \bgroup et al\mbox.\egroup
  }{2001}]{Yu:2001a}
Yu, S.; Lu, J.; Zhu, X.; Duan, H.; Kang, S.; Sun, H.; Wang, H.; Zhao, Q.; and
  Zhan, W.
\newblock 2001.
\newblock Processing norms of modern {Chinese} corpus.
\newblock Technical report, Technical report.

\bibitem[\protect\citeauthoryear{Zhang, Zhang, and
  Fu}{2016}]{zhang2016transition}
Zhang, M.; Zhang, Y.; and Fu, G.
\newblock 2016.
\newblock Transition-based neural word segmentation.
\newblock {\em Proceedings of Annual Meeting of the Association for
  Computational Linguistics, ACL}.

\bibitem[\protect\citeauthoryear{Zheng, Chen, and Xu}{2013}]{zheng2013deep}
Zheng, X.; Chen, H.; and Xu, T.
\newblock 2013.
\newblock Deep learning for chinese word segmentation and pos tagging.
\newblock In {\em Proceedings of the Conference on Empirical Methods in Natural
  Language Processing, EMNLP}.

\bibitem[\protect\citeauthoryear{Zoph and Knight}{2016}]{zoph2016multisourcemt}
Zoph, B., and Knight, K.
\newblock 2016.
\newblock Multi-source neural translation.
\newblock In {\em {NAACL} {HLT} 2016, The 2016 Conference of the North American
  Chapter of the Association for Computational Linguistics: Human Language
  Technologies, San Diego California, USA, June 12-17, 2016}.

\end{thebibliography}
\bibliographystyle{aaai}

\end{document}